%% file: 0_main.tex
\documentclass[letterpaper, 10 pt, conference]{ieeeconf}  

\IEEEoverridecommandlockouts                              

\usepackage[utf8]{inputenc}
\usepackage[T1]{fontenc}
\usepackage{mathrsfs}
\usepackage{amsmath}
\usepackage{amssymb}
\usepackage{amsfonts}
\usepackage{graphicx}
\usepackage{cleveref}
\usepackage{tabstackengine}
\usepackage{mathtools}
\usepackage{multirow}
\usepackage{subcaption}
\usepackage{cite}
\usepackage{algpseudocode}
\usepackage{algorithm}
\usepackage{url}

\graphicspath{ {./images/} }

\title{\LARGE \bf
Trajectory Planning for Autonomous Vehicle Using\\ Iterative Reward Prediction in Reinforcement Learning
}

\author{Hyunwoo Park$^{1}$
\thanks{*This work is supported by the Korea Agency for Infrastructure Technology Advancement(KAIA) grant funded by the Ministry of Land, Infrastructure and Transport.
(RS-2021-KA160853, Road traffic Infrastructure monitoring and emergency recovery support service technology development)}
    \thanks{$^1$ ThorDrive, Seoul, 07268, Republic of Korea}%
}
\begin{document}
\maketitle
\thispagestyle{empty}
\pagestyle{empty}

\begin{abstract}
Traditional trajectory planning methods for autonomous vehicles have several limitations. For example, heuristic and explicit simple rules limit generalizability and hinder complex motions. These limitations can be addressed using reinforcement learning-based trajectory planning. However, reinforcement learning suffers from unstable learning, and existing reinforcement learning-based trajectory planning methods do not consider the uncertainties. Thus, this paper, proposes a reinforcement learning-based trajectory planning method for autonomous vehicles.
The proposed method involves an iterative reward prediction approach that iteratively predicts expectations of future states. These predicted states are then used to forecast rewards and integrated into the learning process to enhance stability. Additionally, a method is proposed that utilizes uncertainty propagation to make the reinforcement learning agent aware of uncertainties.
The proposed method was evaluated using the CARLA simulator. Compared to the baseline methods, the proposed method reduced the collision rate by 60.17\%, and increased the average reward by 30.82 times. A video of the proposed method is available at \url{https://www.youtube.com/watch?v=PfDbaeLfcN4}.

\end{abstract}

\input{1_introduction.tex}
\input{2_related.tex}
\input{3_method.tex}
\input{4_evaluation.tex}

\input{5_result.tex}
\input{6_conclusion_and_future_work.tex}




\bibliographystyle{unsrt}
\bibliography{references}

\end{document}

%% file: 1_introduction.tex
\section{INTRODUCTION}

Since the 2007 DARPA Urban Challenge \cite{buehler2009darpa}, autonomous vehicles(AVs) have been studied intensively. In addition, planning methods for AVs have also been researched intensively and the findings have allowed AVs to drive successfully within limited areas, using rule-based and optimization-based algorithms, explicit heuristic rules, and parameters specified for the given area. However, these traditional approaches suffer from several limitations, including a lack of generality and a lack of complex motion. For example, the heuristic rules and parameters specified for the given area may not be applied to the other areas, which impact the scalability. In addition, many possible scenarios must be considered in real-world applications. If various scenarios are generalized with few scenarios(e.g., lane following, and lane changing), an overly simple policy could be obtained. Numerous studies have employed deep learning to address these limitations \cite{hu2023planning,pini2023safe, shalev2016safe, kendall2019learning, chen2019model, chen2021interpretable, saxena2020driving, wu2023dyna, osinski2020simulation,  gao2022cola, gu2023safe, qiao2021behavior, li2022hierarchical, ma2021reinforcement, phan2023driveirl, tian2021learning}.


 The most popular approach is imitation learning(IL), which learns a driving policy directly from expert driving data \cite{hu2023planning,pini2023safe}; however, IL also has several limitations, including the cost of scalability, simple driving policies, and safety. For example, to scale up the AV using the IL method, expert data must be obtained for all scenarios and targeted areas, which is costly. In terms of driving policies, IL is typically used for handling simpler driving tasks, e.g., lane following. To learn complex policies or policies in corner cases, it should have a lot of data which can be costly and time-consuming. In addition, an expert's demonstrated policies for complex scenarios or corner cases are more distributed than the simple scenarios, which may yield large errors or learning may be infeasible \cite{shalev2016safe}. 
 Relative to safety limitations, insufficient amounts of data on dangerous cases or corner cases are available; thus, the IL agent could output dangerous policies due to a lack of training data.


\begin{figure}[t]
    \centering
    \includegraphics[width=\columnwidth]{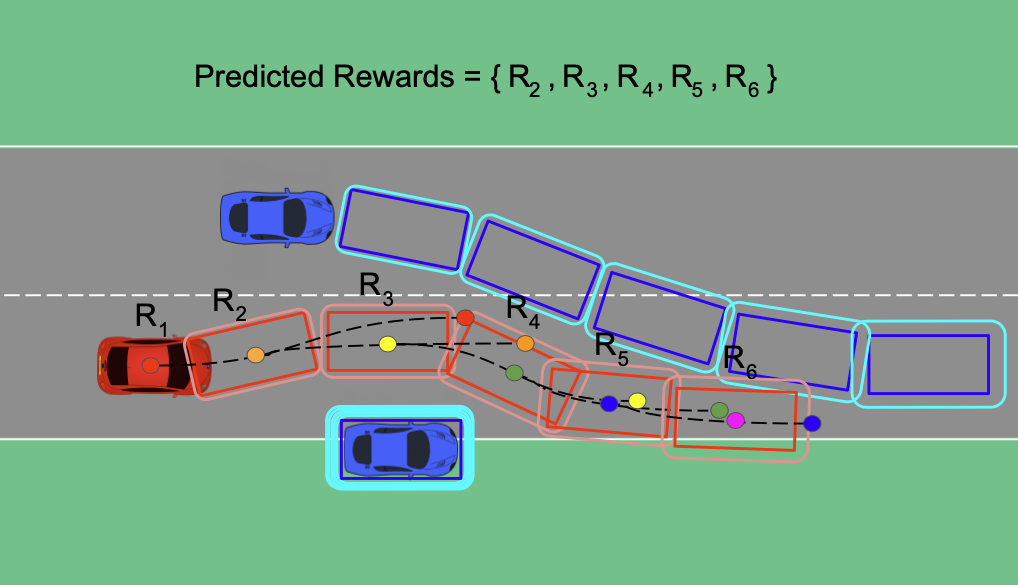}
    \captionsetup{size=footnotesize}
    \caption{The AV(red car) plans a trajectory(red boxes) along its lane while avoiding a parked car(blue car) on the right and a car that is changing lanes(another blue car) from the left lane. The AV's planned trajectories are represented as red boxes, and their trajectories considering uncertainty, are represented as rounded light red boxes. The other 
    vehicles' predicted trajectories are represented as blue boxes, and their trajectories, considering uncertainty, are represented as rounded light blue boxes. The goal of each AV's trajectory is determined iteratively at each time step from the previously predicted state of the AV and the states of the other vehicles. 
    The AV's states and corresponding goals are represented by red, orange, yellow, green, blue, and purple circles(shown in chronological order). 
    The predicted states of the other vehicles and the iteratively predicted states of the AV are employed for the prediction of reward within the planning horizon, which stabilizes the RL learning process.}
    \label{fig:main}
\end{figure}

Another approach is the reinforcement learning(RL) method, which learns a policy via self-exploration and reinforcement without expert data. RL can also simulate and learn both complex policies and policies in corner cases. However, RL suffers from unstable learning when a neural network is used as a function approximation \cite{mnih2015human}. Error of function approximation results in unstable learning or even divergence \cite{van2016deep, fujimoto2018TD3}. In addition, most previous RL-based trajectory planning studies did not consider the uncertainties of the object detection, trajectory prediction of other traffic participants, localization, and control modules that are essential for AV to navigate. Note that not considering uncertainties could cause sudden decelerations or even accidents.

Thus, this study proposes an RL-based trajectory planning method for AVs that overcomes the identified limitations of RL and traditional planning methods. 
The proposed method employs a reward prediction(RP) during the learning process which predicts expectations of future states. These predicted states are then used to forecast rewards and integrated into the learning process to enhance stability. In addition, an iterative RP(IRP) method that uses RP iteratively to predicts relevant states, actions, and corresponding rewards accurately is employed. As a result, the performance of the agent and the learning stability are increased. Additionally, a method is proposed that utilizes uncertainty propagation \cite{xu2014motion, fu2023efficient} to make the reinforcement learning agent aware of uncertainties.
An overview of the proposed method is shown in Fig.\ref{fig:main}.


The primary contributions of this study are summarized as follows:
\begin{itemize}
\item The proposed method increases learning stability and the performance of the RL agent.
\item The proposed method allows the RL agent to be aware of uncertainty.
\item A demonstration and comparison of the proposed method with the baseline methods in the CARLA simulator are presented.
\end{itemize}

The remainder of the paper is organized as follows: Section \ref{related} reviews related works of RL-based trajectory planning methods and uncertainty-aware planning methods.
Section \ref{method} defines problem formulation and RP, IRP, and application of uncertainty propagation is proposed. Section \ref{evaluation} shows how the proposed method and baseline methods are evaluated in the CARLA simulator. Section \ref{result} analyzes the evaluation results and shows how the key metrics are improved. Section \ref{conclusion_and_future_work} concludes the proposed method and discusses future works.

%% file: 2_related.tex
\section{Related Works} \label{related}

\subsection{RL-based Trajectory Planning}

Previous RL-based trajectory planning methods for AVs can be divided into two categories according to the action of an RL agent, i.e., 1) control command and 2) the Goal of the trajectory. 

\subsubsection{Control Command} 
Methods that action of an RL agent is a control command \cite{kendall2019learning, saxena2020driving, chen2019model, chen2021interpretable, osinski2020simulation, wu2023dyna, li2022hierarchical} use a lateral control(e.g., steer angle and steering rate) and a longitudinal control(e.g., acceleration and jerk) as an output of an RL agent. However, such methods tend to fail to learn easily. The variance of action affects the learning process; thus, to drive an AV successfully using a control command, very specific policies are required to yield good rewards. For example, in a highway scenario, a small turn of the steering wheel may yield catastrophic results. This specific policy requirement makes it difficult for RL agents to explore and find good states and actions, which leads to sample inefficiency during training and ineffective learning. Thus, unless the RL agent find good policy early on by chance, learning will fail. In addition, an agent's intentions are unknown; thus they lack interpretability.

Kendall et al. \cite{kendall2019learning} employed monocular images as an observation and DDPG as the main algorithm to follow the lane in real-world scenarios. In addition, Chen et al. \cite{chen2019model}, employed the bird-view semantic mask as an observation and evaluated their method in CARLA simulator. Their work was developed further \cite{chen2021interpretable} by increasing interpretability using the probabilistic graphical model. Saxena et al. \cite{saxena2020driving}, employed a field-of-view as an observation and proximal policy optimization(PPO) as the main algorithm. Their primary task was lane changing in dense traffic scenarios. In addition, Wu et al. \cite{wu2023dyna}, employed the Dyna algorithm with PPO as the main algorithm, and they imitated the world model using the Gaussian process. Li et al. \cite{li2022hierarchical}. proposed a method using an hierarchical RL(HRL). Their model-based high-level policy generates subgoals via optimization that utilizes a low-level policy and an offline low-level policy outputs a control command.

 
\subsubsection{Goal for Trajectory}
Approaches that action of RL agent is goal/goals for trajectory \cite{qiao2021behavior, gao2022cola, gu2023safe, ma2021reinforcement} are comparably stable in learning. The action variance of these approaches has a relatively small effect on learning process of RL compared to approaches that action of an RL agent is control command. For example, in a highway scenario, a small change in the lateral deviation of a goal would result in a smaller amount of change in a result compared to the result of the control command example.

Gao et al. \cite{gao2022cola} proposed method using HRL. Their high-level policy generates subgoals in the Frenet frame to guarantee the temporal and spatial reachability of the generated goal and the low-level policy outputs control commands. Their work was developed further \cite{gu2023safe} by ensuring safety using the safe-state enhancement method. In addition, Qiao et al. \cite{qiao2021behavior} employed a hybrid HRL method, where a high-level policy generates optimal behavior decisions, and a low-level policy generates a trajectory point that the AV intends to trace. They also employed a PID controller to trace the trajectory point. Ma et al. \cite{ma2021reinforcement} used the latent state inference method, and employs PPO as a main algorithm.

\subsection{Uncertainty-aware Planning}
Uncertainty-aware planning methods are used to plan a trajectory for an AV by considering the uncertainty of the AV(i.e., localization and control) and the traffic participants(i.e., object detection, and trajectory prediction). Xu et al. \cite{xu2014motion} employed a Kalman filter to estimate the uncertainty of the traffic participants, and the LQG framework to estimate the uncertainty of the AV. They also used the uncertainty estimation in planning by widening the size of the AV and the traffic participants when checking for the collision conditions. Fu et al. \cite{fu2023efficient} and Qie et al. \cite{qie2022improved} also employed a Kalman filter to estimate the uncertainty of the traffic participant. Fu et al. used estimated uncertainty as a chance constraint when planning a velocity profile, and Qie et al. employed estimated uncertainty in a tube-based MPC to plan a trajectory. In addition, Hubmann et al. \cite{hubmann2018automated} formulated the planning problem with uncertainties as a partially observable Markov decision process. They estimated the intent of a traffic participant and utilized it as an uncertainty. By using the adaptive belief tree and uncertainty,
 they determined the optimal longitudinal motion of the AV. Khaitan et al. \cite{khaitan2021safe} estimated the uncertainty of the traffic participants by utilizing reachable set in short-term horizon and used it in the tube MPC to execute the trajectory safely in the presence of uncertainty. 


%% file: 3_method.tex
\section{Method} \label{method}
 To solve trajectory planning problems using RL, RL algorithms with continuous action space, e.g., the DDPG \cite{lillicrap2015DDPG}, TD3 \cite{fujimoto2018TD3}, and PPO \cite{schulman2017proximal} algorithms, are more suitable than algorithms that utilize a discrete action space. Since getting smooth behaviors requires an increase in the size of the discrete action space which leads to discrete control methods being intractable. 
 In addition, for simplicity and interpretability, algorithms with deterministic policies(rather than stochastic policies) are selected.
 Furthermore, generating the goal for a trajectory is a better action choice than control command because more specific policies are required for control command methods to yield rewards successfully. Thus, the proposed trajectory planning method generates a goal using a deterministic continuous control RL algorithm. It is assumed that information about the localization, route path, trajectory prediction of other traffic participants, and object detection is provided. However, trajectory prediction of other traffic participants is only used during the learning process.

\subsection{Problem Formulation} \label{sec:problem_formulation}
The trajectory planning problem is formulated as a Markov decision process, which is defined by the tuple $(S,A,P,R)$. Here, $s \in S$ is the continuous state space, $a \in A$ is the continuous action space, $P$ is the probability of state transition, and $R$ is the reward received after each transition. The return is defined as discounted sum of rewards $G_t = \sum_{k=0}^{\infty} \gamma^{k}R_{t+k+1}$, where $\gamma \in (0, 1)$ is the discount factor determining the priority of short-term rewards. The purpose of RL is to learn an optimal policy that maximizes the expected cumulative reward as follows:

\begin{equation} \label{eq:rp_1}
\begin{split}
     \max_{\pi}J(\pi) = \mathbb{E}_{s\sim\rho^{\pi}, a\sim\pi}[\sum_{i=0}^{\infty} \gamma^{i}r(s,a)],
\end{split}
\end{equation}
where $r$ is the reward function, $\rho^\pi$ is the state distribution under the policy $\pi$. 
The future states of the AV and other traffic participants are assumed to follow the planned trajectory and the predicted trajectory, respectively, with deviations following a normal distribution, as assumed in \cite{xu2014motion}.

\subsection{Reward Prediction} \label{sec:reward_prediction}
Continuous control RL algorithms employ the policy gradient method to learn policies directly. The policy gradient method maximizes the following objective function.
\begin{equation} \label{eq:rp_2}
\begin{split}
    \nabla_{\theta}J(\pi_{\theta}) = \mathbb{E}_{s\sim\rho^{\pi}, a\sim\pi_{\theta}}[\nabla_{\theta}log\pi_{\theta}(a|s)Q^{\pi}(s,a)].
\end{split}
\end{equation}

This theorem is derived from the following objective function:
\begin{equation} \label{eq:rp_3}
\begin{split}
    J(\pi_{\theta}) = \mathbb{E}_{s\sim\rho^{\pi}, a\sim\pi_{\theta}}[r(s,a)].
\end{split}
\end{equation}

The objective function $J(\pi_{\theta})$ in (\ref{eq:rp_2}) can be defined as the action value function $Q$, the advantage function $A$, and the TD error $\delta$. The action value and advantage functions are approximated with neural networks in the deep RL method; however, function approximation with neural network always involves errors, which causes instability during the learning process and poor performance. To address this, the N-step SARSA concept is employed in RP. N-step SARSA improves learning stability by utilizing the error reduction property of n-step returns. Nevertheless, the learning process of N-step SARSA remains susceptible to instability.

To overcome this, the RP method that utilizes the error reduction property of the n-step returns and enhances the learning stability is proposed. To achieve this stability, RP enables the agent to explicitly learn the consequences of actions. This is accomplished by utilizing predictions of traffic participants' behavior and the planned trajectory of the AV, which is obtained using the agent's output goal. Specifically, future rewards are predicted by considering the planned future states of the AV and the predicted states of other traffic participants. Thus, RP leverages the advantages of N-step SARSA while explicitly learning action consequences. These two traits of RP contribute to stabilizing the learning process.
The proposed RP method is employed during the action value function update process as follows:


\begin{figure}[t]
    \centering
    \includegraphics[width=\columnwidth]{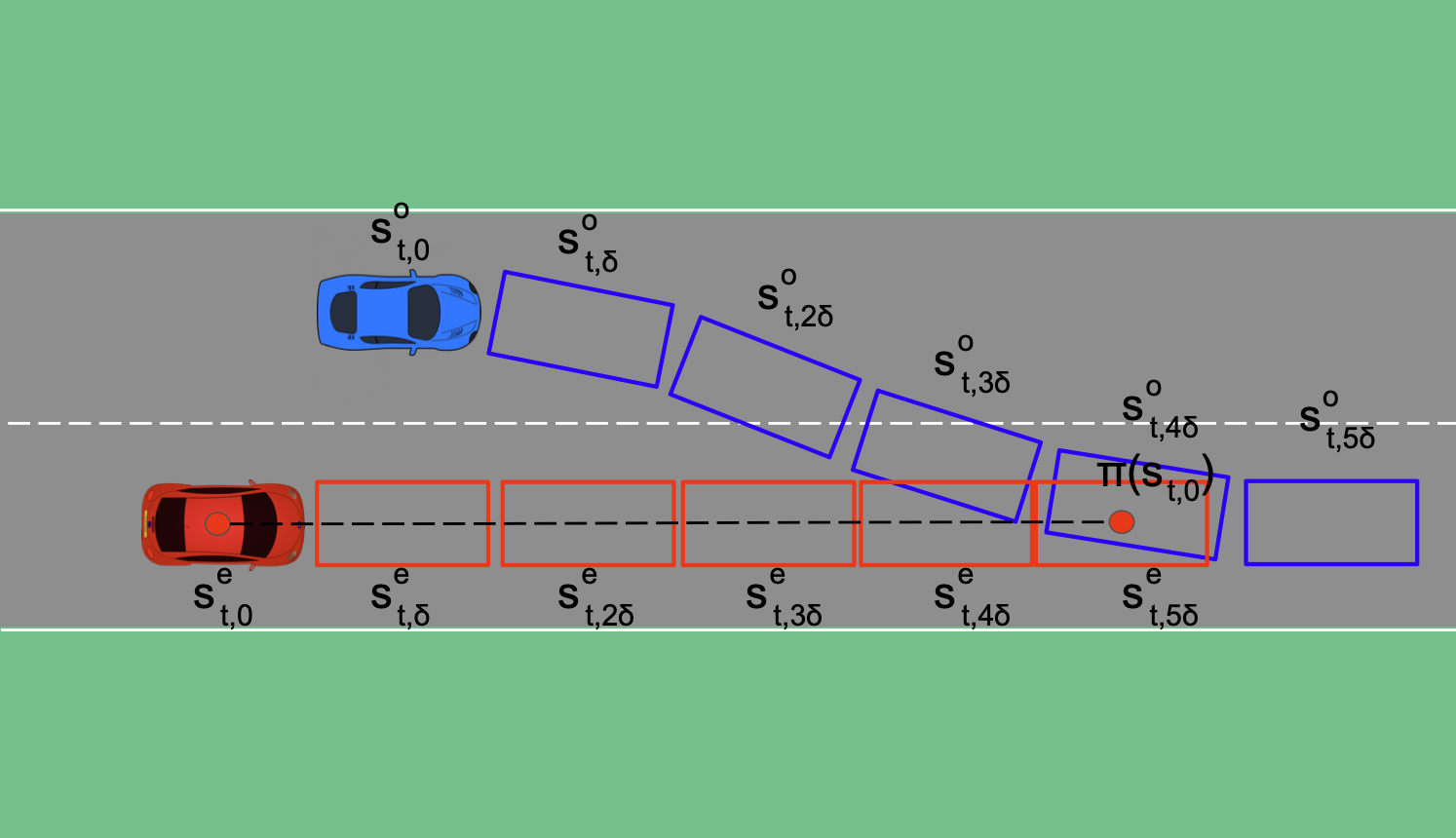}
    \captionsetup{size=footnotesize}
    \caption{The AV(red car) is following the lane while the other traffic participant(blue car) attempts to change lanes. The predicted states of the AV and the other traffic participant are represented as red and blue boxes respectively. The state of the AV at time $t$ and its goal is represented as red circles. The predicted states of the other traffic participant are assumed to be given. The predicted states(red boxes) of the AV are predicted by the RL agent's action $\pi(s_{t,0})$, i.e., the goal of the trajectory.}
    \label{fig:RP}
\end{figure}

\begin{equation} \label{eq:rp_4}
\begin{split}
    &s^e_{t,\tau} \in \mathcal{T}^e_t, \; s^{o,k}_{t,\tau} \in \mathcal{T}^{o,k}_t, \; s^o_{t,\tau} = \{s^{o,}_{t,\tau}, s^{o,1}_{t,\tau}, \cdots s^{o,n}_{t,\tau}  \} \; \\
    &s_{t,\tau} = f(s^e_{t,\tau},s^o_{t,\tau}), \; R_{t,\tau+\delta} = g(s_{t,\tau},s_{t,\tau+\delta}),  \\
    &J(\pi)= Q^{\pi}(s_t,a_t),\\
    &L(\theta^Q) = \mathbb{E}_{s_t\sim\rho^{\pi}, a_t\sim\pi_{\theta}}[(Q^{\pi}(s_t,a_t) - y_t)^2],\\
    &y_t = R_{t,\delta} + \gamma \cdot R_{t, 2\delta} +\cdots \\
    & \quad \quad \quad + \gamma^{T/ \delta - 1} \cdot R_{t,T} + \gamma^{T/ \delta} \cdot Q^{\pi}(s_{t,T},\pi(s_{t,T})), \\
\end{split}
\end{equation}

where time $t$ is when the AV's trajectory is planned and the trajectory prediction of other traffic participants is made, $T$ is a planning/prediction horizon, $\delta$ is a sufficiently small time step size, 
$\tau$ is the time within the planning horizon $\tau \in [\delta,T]$, $s^e_{t,\tau}$ is the predicted future state of the AV at time $t + \tau$ from trajectory $\mathcal{T}^e_t$ planned at time $t$ using the goal $\pi(s_{t,0})$,
$s^{o,k}_{t,\tau}$ is the predicted future state of the $k$th other traffic participant at time $t + \tau$ from prediction $\mathcal{T}^{o,k}_t$ predicted at time $t$,
$s^{o}_{t,\tau}$ is the predicted future states of $n$ other traffic participants at time $t + \tau$ from prediction at time $t$. In addition, $f$ is a function that combines the predicted future states of the AV and the other traffic participants at the same time to output the state $s_{t,\tau}$ for the RL agent, 
$g$ is a function that predicts the reward $R_{t,\tau+\delta}$ at time $t+\tau+\delta$ during transition from $s_{t,\tau}$ to $s_{t,\tau+\delta}$, 
$L(\theta^Q)$ is the loss for the action value function $Q^{\pi}$, $y_t$ is the expected return inferred by utilizing predicted rewards and the action value of the predicted state at the planning horizon, and its corresponding action, and $R_{t,\delta}$ is a reward received from the environment. Predicted future states are used to predict rewards, which are then utilized during the action value function update process. 
Fig.\ref{fig:RP} shows the predicted states of the other vehicle and the AV with its goal at time $t$.

Unlike N-step SARSA, the expected return from the action value function learned by RP is not based on rewards resulting directly from state transitions, but rather on rewards associated with expected states. Namely, RP increases learning stability by utilizing expected future states, yet it does not directly account for the variation of these future states. However, in terms of predicting the states of other traffic participants and the AV, its variance is strongly related to safety. Therefore, uncertainty propagation on future states of other traffic participants and the AV is utilized to consider this variance. Additional details about the uncertainty propagation process are given in Section \ref{sec:uncertainty_propagation}.

\begin{figure}[t]
    \centering
    \includegraphics[width=\linewidth]{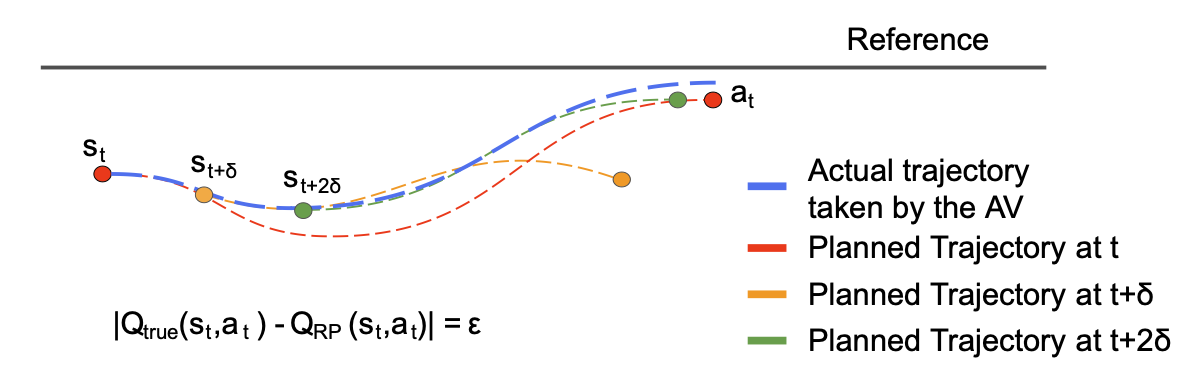}
    \captionsetup{size=footnotesize}
    \caption{Demonstration of inaccuracy problem of RP. Here, the agent has planned a trajectory that takes AV close to the reference. The trajectories planned at each time step beginning from each state-action pair are represented by different colors. The return predicted using RP, $Q_{RP}$, is based on the red trajectory planned at time $t$. The agent plans a new trajectory at each step; therefore, the planned trajectory that RP utilized at time $t$(red) and the actual trajectory executed by the AV(blue) are different. Thus, $Q_{RP}$ has an error $\epsilon$ compared to the true return $Q_{true}$, which is the return obtained by following the blue trajectory.}
    \label{fig:IRP_example}
\end{figure}

\subsection{Iterative Reward Prediction} \label{sec:iterative_reward_prediction}
In general, AVs plan a trajectory at every time step, which is a common practice due to potential inaccuracies in predicting the behavior of other traffic participants and the imprecise path tracking by the controller. Therefore, the proposed method also assumes trajectory planning at every time step. Given this assumption, even if the controller tracks the given trajectory perfectly, there is no guarantee that the trajectory generated using the agent's output will be the same as the previous one. Consequently, this inaccuracy of RP leads to inaccurate prediction of rewards, resulting in learning instability and poor performance. Fig.\ref{fig:IRP_example} demonstrates the inaccuracy problem of RP. 

\begin{figure}[t]
    \centering
    \includegraphics[width=\columnwidth]{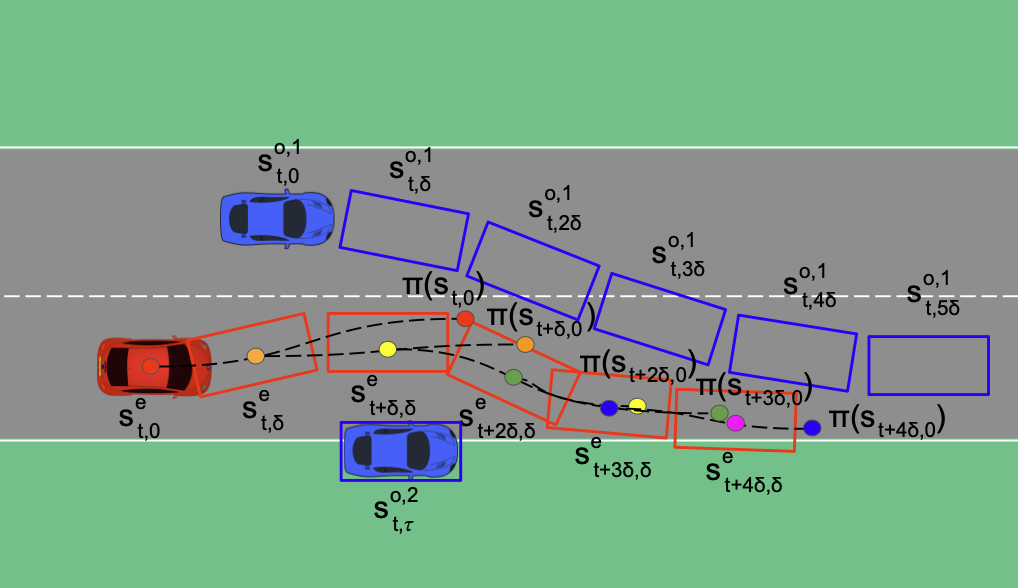}
    \captionsetup{size=footnotesize}
    \caption{Predicted states(blue boxes) of other traffic participants(blue cars) and predicted states(red boxes) of the AV(red car) and actions of the RL agent. The action of the RL agent is determined iteratively from the previous state of the AV, its goal, and the states of the other vehicles. The goals and the AV's start states are represented as red, orange, yellow, green, blue, and purple circles in chronological order. The predicted states of the other vehicles and the iteratively predicted states of the AV are utilized to predict the rewards during the learning process.}
    \label{fig:IRP}
\end{figure}

To address the inaccuracy problem of RP, the IRP method is proposed. This method predicts the rewards by iteratively planning a new trajectory at the predicted state and predicting the reward of that trajectory for one time step. Fig. \ref{fig:IRP} shows how the proposed IRP method operates. Compared to the conventional RP, the proposed IRP predicts the rewards more accurately and reduces the error of the function approximation of the action value function $Q^{\pi}$. The mathematical representation of the IRP is given as follows:


\begin{equation} \label{eq:irp}
\begin{split}
    &s_{t+\tau,0} = f(s^e_{t+\tau,0},s^o_{t,\tau})\\ &s_{t+\tau,\delta} = h(s_{t+\tau,0},\pi(s_{t+\tau,0})),\\
    &R_{t+\tau,\delta}= g(s_{t+\tau,0}, s_{t+\tau,\delta}),\\
    &J(\pi)= Q^{\pi}(s_t,a_t),\\
    &L(\theta^Q) = \mathbb{E}_{s_t\sim\rho^{\pi}, a_t\sim\pi}[(Q^{\pi}(s_t,a_t) - y_t)^2],\\
    &y_t = R_{t,\delta} + \gamma \cdot R_{t+\delta, \delta} +\cdots + \gamma^{T/ \delta - 1} \cdot R_{t+T-\delta,\delta}\\
    & \quad \quad \quad + \gamma^{T/ \delta} \cdot Q^{\pi}(s_{t+T-\delta,\delta},\pi(s_{t+T-\delta,\delta})), \\
\end{split}
\end{equation}


where time $t$ is when planning first started, 
$s_{t+\tau,\delta}$ is the predicted state at time $t+\tau+\delta$ predicted from the previously predicted state at time $t+\tau$, and 
$h$ is a function that outputs a predicted state using the current state, and  action. 
Here, the predicted state, such as  $s_{t+\tau,\delta}$ is used as
$s_{t+\tau+\delta,0}$ for iteratively predicting the next state $s_{t+\tau+\delta,\delta}$. In function $h$, the future states of the other traffic participants are assumed to follow the trajectories predicted at time $t$, which is assumed to be given. However, the future states of the AV are updated iteratively at each time step. 

In addition, considering the agent's future action more accurately enables the agent to drive along a complicated trajectory because the agent has a more comprehensive understanding of its future actions. As shown in Fig.\ref{fig:IRP}, the AV can plan a red trajectory by considering the agent's future actions; thus, it is capable of driving a complicated trajectory.

\subsection{Uncertainty Propagation} \label{sec:uncertainty_propagation}
 During RP and IRP, the expected future states are considered; however, the variance and uncertainty of the predicted future states are not considered. Note that not considering the variance of the prediction leads to inaccurate function approximation and the RL agent being unaware of the uncertainty. In addition, ignoring the uncertainty can cause the agent to perform sudden deceleration or become involved in an accident. For example, if a low-functioning controller is employed to track a trajectory, and the RL agent does not take that into account, then the AV may encounter various dangerous situations and the predicted rewards may have significant error. To consider the variance of the prediction and the uncertainty of future states, uncertainty propagation on the RP and IRP motivated by \cite{xu2014motion,fu2023efficient} is utilized. 

The uncertainty propagation process is built on the Kalman filter; however, the measurement update process of the Kalman filter is removed because observing future states is impossible. The uncertainty propagation utilized on the RP is modeled as follows:

\begin{equation} \label{eq:up_1}
\begin{split}
    &\check{s}^e_{t,\tau} \sim N(s^e_{t,\tau}, \Sigma^e_{t,\tau}), \;
    \check{s}^o_{t,\tau} \sim N(s^o_{t,\tau}, \Sigma^o_{t,\tau}) \\
    &\Sigma^e_{t,\tau+1} = F \Sigma^e_{t,\tau} F^T + Q^e_\tau, \;
    \Sigma^o_{t,\tau+1} = F \Sigma^o_{t,\tau} F^T + Q^o_\tau, \;
\end{split}
\end{equation}
where each state of the AV $\check{s}^e_{t,\tau}$, and states of the other traffic participants $\check{s}^o_{t,\tau}$ is modeled as Gaussian a random variable with means $s^e_{t,\tau}$, and $s^o_{t,\tau}$, and covariance of $\Sigma^e_{t,\tau}$, and $\Sigma^o_{t,\tau}$, respectively, based on the assumption made on \ref{sec:problem_formulation}, 
$F$ is the state transition matrix and $Q^e_\tau$, $Q^o_\tau$ represent the process noise of the AV and the other traffic participants, respectively, where $Q^e_\tau$ is attributed to localization, control error, and  $Q^o_\tau$ is attributed to object detection, trajectory prediction error respectively. 
The introduced states $\check{s}^e_{t,\tau}$ and $\check{s}^o_{t,\tau}$ are utilized in the collision checking process of the IRP method to account for uncertainty.
Here, the ellipse defined by the covariance matrix can provide an upper bound of the probability $1 - \delta$ that the AV and other traffic participants exist. However, to check for a collision between the AV and other traffic participants, the rectangle shapes of the AV and other traffic participants must be considered. In the proposed method, the Minkowski sum of the rectangle and the ellipse \cite{xu2014motion} is computed. The new shape from the Minkowski sum is then utilized for collision checking, which guarantees the probability of $(1- \delta)^2$ whether collide or not. The illustration of uncertainty propagation using the Minkowski sum is shown in Fig.\ref{fig:main}.

\subsection{Overall Algorithm} \label{sec:overall_algorithm}
Below, a demonstration showcasing the combination of IRP and uncertainty propagation is provided. 
The proposed method, based on the existing deterministic policy gradient algorithm, is applied during the critic update process.
The pseudocode of the proposed method is given in Algorithm \ref{alg:critic}. Here, in lines 6-7, the uncertainty propagation and Minkowski sum are executed using the state of the AV and the other traffic participants $s^e_{t+\tau,0},s^o_{t+\tau,0}$. In line 9, the state $s_{t+\tau,\delta}$ results from the utilization of the agent's previous state $s_{t+\tau,0}$, and the agent's action $\pi(s_{t+\tau,0})$.
In line 10, the reward is predicted during the transition from $s_{t+\tau,0}$ to $s_{t+\tau,\delta}$. In line 12, the state $s_{t+\tau,\delta}$ is divided into $s^e_{t+\tau,\delta}$, and $s^o_{t+\tau,\delta}$, which are used as the states $s^e_{t+\tau+\delta,0}$, and $s^o_{t+\tau+\delta,0}$ for the next iteration.


\begin{algorithm}
\caption{Pseudo code of proposed method}\label{alg:critic}
\begin{algorithmic}[1]
    \Procedure{CriticUpdate}{ }
    \State $s^e \gets s^e_{t,0}, s^o \gets s^o_{t,0}$
    \State $r \gets R_{t,\delta}$ \Comment{Initialize predicted return with received reward from the simulator}
    \State $\tau \gets \delta$
    \While{$\tau < T$}
        \State $\check{s}^e, \check{s}^o \gets UncertaintyPropagation(s^e, s^o)$
        \State ${s^e}',{s^o}' \gets MinkowskiSum(\check{s}^e, \check{s}^o)$
        \State $s \gets f({s^e}',{s^o}')$ \Comment{Merge states}
        \State $s' \gets h(s,\pi(s))$ \Comment{Prediction of $s_{t+\tau,\delta}$}
        \State $R  \gets g(s,s')$ \Comment{Predict the reward $R_{t+\tau,\delta}$}
        \State $r \gets r + \gamma^{\tau/\delta}R$ \Comment{Update the predicted return}
        \State $s^e,s^o \gets f^{-1}(s')$ \Comment{Update the next state}
        \State $\tau \gets \tau + \delta$
    \EndWhile
    \State $\check{s}^e, \check{s}^o \gets UncertaintyPropagation(s^e, s^o)$
    \State ${s^e}',{s^o}' \gets MinkowskiSum(\check{s}^e, \check{s}^o)$
    \State $s \gets f({s^e}',{s^o}')$ \Comment{Merge states}
    \State Set $y_t = r + \gamma^{T/\delta}Q(s,\pi(s|\theta^\pi)|\theta^Q)$
    \State Update critic by minimizing the loss: 
    \State $L(\theta^Q) =(Q(s_{t},a_{t}|\theta^Q))-y_t)^2$
    \EndProcedure
\end{algorithmic}
\end{algorithm}

%% file: 4_evaluation.tex
\section{Experiments} \label{evaluation}
The proposed method and baseline methods were evaluated using the CARLA simulator. The experimental configuration, baseline methods, and implementation details are described in the following sections.
\subsection{Experiment Configuration}
The proposed and baseline methods were evaluated in four distinct scenarios. Scenario 1 involved lane following with static obstacles, scenario 2 involved lane following with traffic participants, scenario 3 involved lane changing with traffic participants, and scenario 4 entailed overtaking parked cars with traffic participants. All necessary inputs, e.g., route path, object detection, trajectory prediction, and localization were given. Success was determined if the AV reached the goal without a collision within specified time. In this evaluation, the goal was $130m$ ahead of the AV's initial position. Here, a maximum lateral deviation of $1.5m$ from the center of the target lane was permitted.

In each scenario, the AV was spawned on the road with a random lateral deviation of $[-1.5m, 1.5m]$ from the center of the road, and a random heading angle deviation of $[-20deg, 20deg]$ with a random initial speed of $[5km/h, 15km/h]$.
In scenario 1, a maximum of two static obstacles(i.e., vehicles) were spawned at random positions with a lateral deviation of $[-0.5m,0.5m]$ from the center of the road, and a heading angle deviation of $[-20deg, 20deg]$. Scenario 2 included a maximum of five randomly spawned traffic participants. In scenario 3, the other traffic participants were the same as in scenario 2; however, the AV's goal was to change lanes. In scenario 4, a maximum of two static obstacles(parked cars) and three traffic participants were spawned, and the goal of the AV was to overtake the parked cars while avoiding collisions with the other traffic participants. Note that the traffic participants in each scenario were designed to change lanes randomly.

\subsection{Baseline Methods}
The following two baseline methods were considered in this evaluation. In baseline 1, the output of an agent was a control command, and it is identical to the method proposed in \cite{kendall2019learning}, except that in \cite{kendall2019learning}, monocular images are used as input.
In baseline 2, the output of the agent was the trajectory goal, the same as that of the proposed method, for fair comparison.
However, baseline 2 did not include RP, IRP, or uncertainty propagation.
In addition, proposed method was evaluated individually as follows: RP, IRP, and IRP with uncertainty propagation. 

\subsection{Implementation Details}
All five methods, i.e., baseline1, baseline2, RP, IRP, and IRP with uncertainty propagation were implemented using the DDPG algorithm with the same state, and reward function. 
\subsubsection{State}
The features of the state space $s$ are composed of $s^e$ and $s^o$. Here, $s^e$ is the state of the AV and comprises $(d, \dot{d}, \ddot{d}, \dot{\sigma}, \ddot{\sigma}, \theta, v_{speed \_ limit}, l, w)$, where $\sigma$ and $d$ are the longitudinal and lateral position on the Frenet frame, $\theta$ is the heading angle difference with the center of the road, $v_{speed \_ limit}$ is the speed limit of the road, and $l$ and $w$ are length and width of the AV in consideration of the Minkowski sum. $s^o$ is composed of $s^{o,k}, k \in \mathbb{N}$,
where $\mathbb{N}$ is a natural number. $s^{o,k}$ is the state of the $k$th traffic participant, which is composed of $(\sigma,d,\theta,l,w,v_\sigma,v_d)$, where $v_\sigma$ and $v_d$ are the longitudinal and lateral velocity on the Frenet frame.



\subsubsection{Action}
The action space $a$ is the goal of the trajectory which is composed of $(T_{target}, d_{target}, \sigma_{target},\dot{\sigma}_{target})$, where $T_{target}$ is the time interval between the current state and the goal state, $\sigma_{target}$, and $d_{target}$ are the longitudinal and lateral target positions on the Frenet frame, and $\dot{\sigma}_{target}$ is the target longitudinal speed. Here, the trajectory planning method \cite{werling2010optimal} is employed to plan a trajectory toward the goal. The lateral jerk-optimal trajectory is generated given the initial state of the AV $[d,\dot{d},\ddot{d}]$, and the end state $[d_{target}, \dot{d}_{target}=0, \ddot{d}_{target}=0]$ at $T_{target}$ from the action. In addition, the longitudinal jerk-optimal trajectory is generated given the initial state of the AV $[\sigma,\dot{\sigma},\ddot{\sigma}]$, and the end state $[\sigma_{target}, \dot{\sigma}_{target}, \ddot{\sigma}_{target}=0]$ at $T_{target}$ from the action. The final trajectory is obtained by combining the lateral and longitudinal trajectories. Note that the planned trajectory is tracked using an MPC-based controller.

\subsubsection{Reward}
The reward function is designed to encourage safe, comfortable, and efficient driving as follows:
\begin{equation} \label{eq:reward}
\begin{split}
    R = &\lambda_{lat \textunderscore acc} \cdot a_{lat \textunderscore acc} + \lambda_{lat \textunderscore jerk} \cdot a_{lat \textunderscore jerk} \\
    + &\lambda_{long \textunderscore acc} \cdot a_{long \textunderscore acc} + \lambda_{long \textunderscore jerk} \cdot a_{long \textunderscore jerk} \\
    +&\lambda_{d} \cdot |d| + \lambda_{v} \cdot |v-v_{des}| + r_{collision},
\end{split}
\end{equation}
where $\lambda_{lat \textunderscore acc}$, and $a_{lat \textunderscore acc}$ represent the weight and penalty for the lateral acceleration, $\lambda_{lat \textunderscore jerk}$, and $a_{lat \textunderscore jerk}$ are the weight and penalty for the lateral jerk, $\lambda_{long \textunderscore acc}$, and $a_{long \textunderscore acc}$ are the weight and penalty for the longitudinal acceleration, $\lambda_{long \textunderscore jerk}$, and $a_{long \textunderscore jerk}$ are the weight and penalty for the longitudinal jerk, $\lambda_{d}$, and $|d|$ are the weight and penalty for the lateral deviation from the target lane, $\lambda_{v}$, $|v-v_{des}|$ are the weight and the penalty for being slower or faster than the desired speed respectively, and $r_{collision}$ represents the reward and penalty for a collision event. Here, $r_{collision}$ is negative when a collision occurs and positive when no collision occurs. Note that the above reward is also utilized during RP. 


%% file: 5_result.tex
\section{Results} \label{result}

\begin{figure*} [t!]
    \centering
    \begin{subfigure}[b]{0.32\textwidth}        
        \centering
        \includegraphics[width=\textwidth]{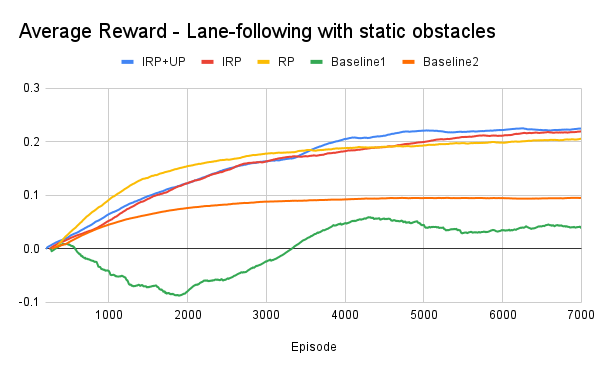}
        \caption{}
        \label{fig:Average_Reward_Obstacles}
    \end{subfigure}
    \hfill
    \begin{subfigure}[b]{0.32\textwidth}        
        \centering
        \includegraphics[width=\textwidth]{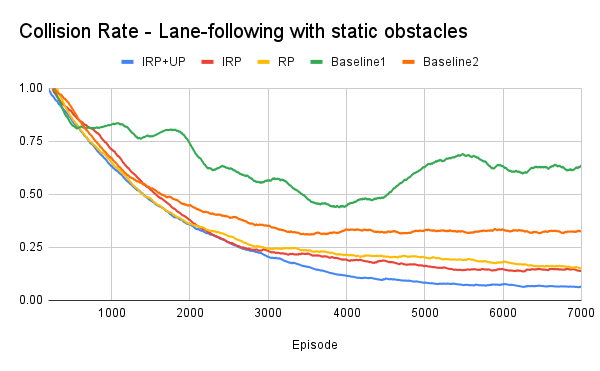}
        \caption{}
        \label{fig:Collision_Rate_Obstacles}
    \end{subfigure}
    \hfill
    \begin{subfigure}[b]{0.32\textwidth}        
        \centering
        \includegraphics[width=\textwidth]{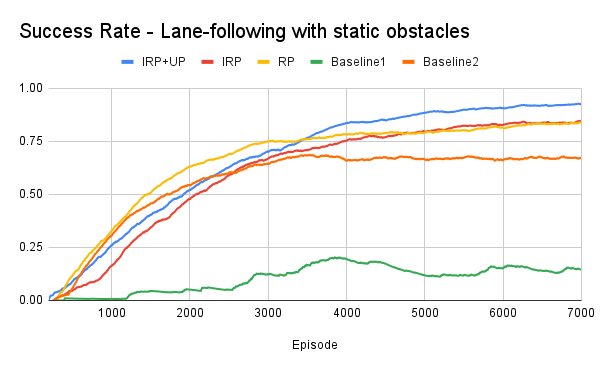}
        \caption{}
        \label{fig:Success_Rate_Obstacles}
    \end{subfigure}

    \begin{subfigure}[b]{0.32\textwidth}        
        \centering
        \includegraphics[width=\textwidth]{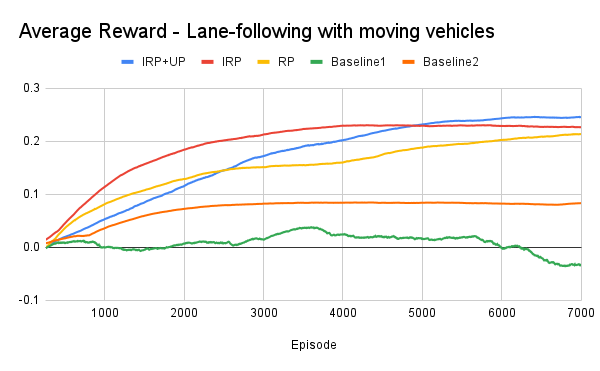}
        \caption{}
        \label{fig:Average_Reward_Vehicles}
    \end{subfigure}
    \hfill
    \begin{subfigure}[b]{0.32\textwidth}        
        \centering
        \includegraphics[width=\textwidth]{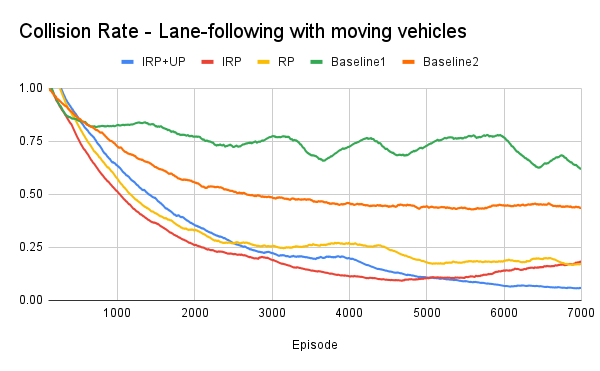}
        \caption{}
        \label{fig:Collision_Rate_Vehicles}
    \end{subfigure}
    \hfill
    \begin{subfigure}[b]{0.32\textwidth}        
        \centering
        \includegraphics[width=\textwidth]{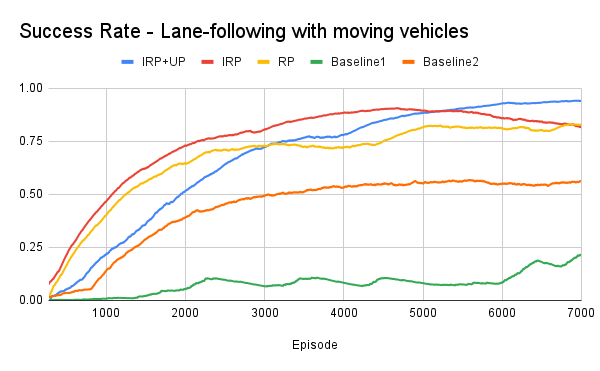}
        \caption{}
        \label{fig:Success_Rate_Vehicles}
    \end{subfigure}

    \begin{subfigure}[b]{0.32\textwidth}        
        \centering
        \includegraphics[width=\textwidth]{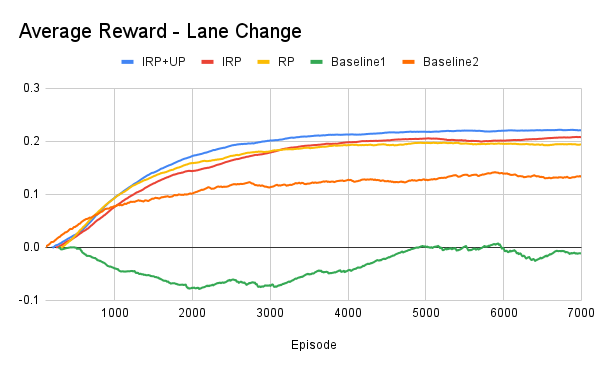}
        \caption{}
        \label{fig:Average_Reward_Lanechange}
    \end{subfigure}
    \hfill
    \begin{subfigure}[b]{0.32\textwidth}        
        \centering
        \includegraphics[width=\textwidth]{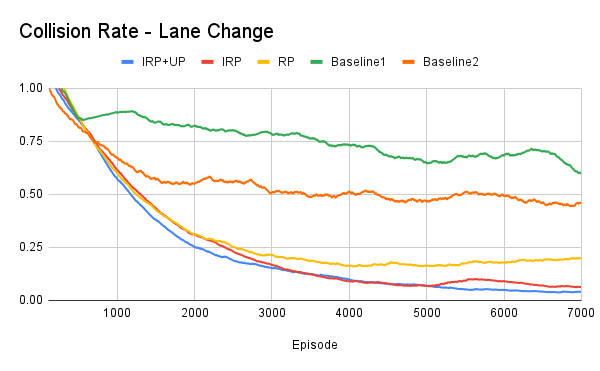}
        \caption{}
        \label{fig:Collision_Rate_Lanechange}
    \end{subfigure}
    \hfill
    \begin{subfigure}[b]{0.32\textwidth}        
        \centering
        \includegraphics[width=\textwidth]{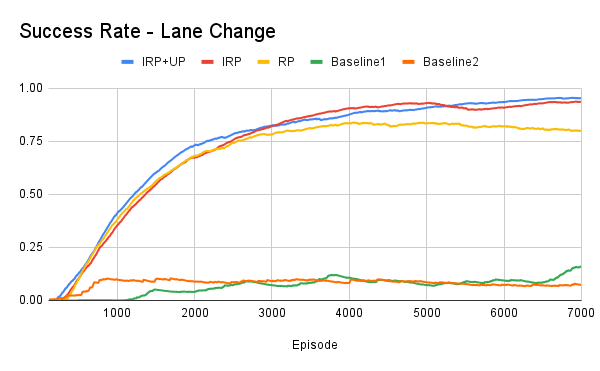}
        \caption{}
        \label{fig:Success_Rate_Lanechange}
    \end{subfigure}

        \begin{subfigure}[b]{0.32\textwidth}        
        \centering
        \includegraphics[width=\textwidth]{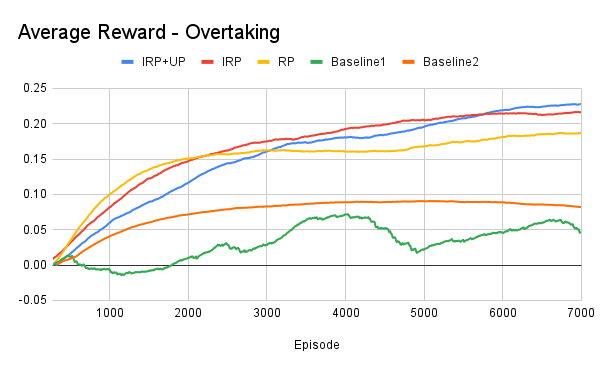}
        \caption{}
        \label{fig:Average_Reward_Detour}
    \end{subfigure}
    \hfill
    \begin{subfigure}[b]{0.32\textwidth}        
        \centering
        \includegraphics[width=\textwidth]{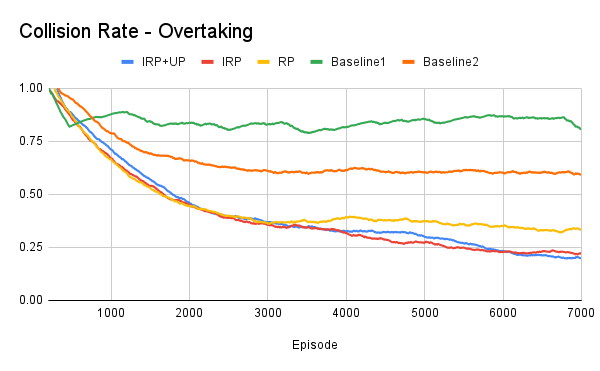}
        \caption{}
        \label{fig:Collision_Rate_Detour}
    \end{subfigure}
    \hfill
    \begin{subfigure}[b]{0.32\textwidth}        
        \centering
        \includegraphics[width=\textwidth]{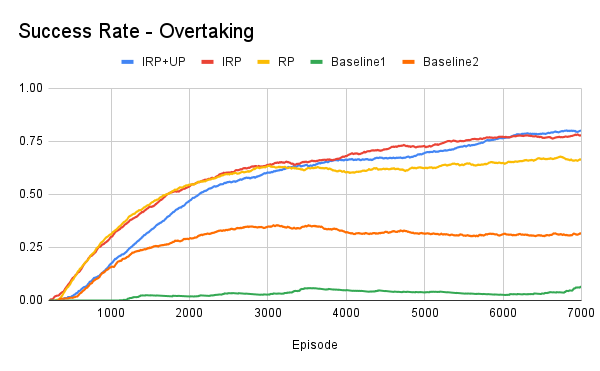}
        \caption{}
        \label{fig:Success_Rate_Detour}
    \end{subfigure}

    \captionsetup{size=footnotesize}
    \caption{Average reward per time step in an episode, collision rate, and success rate of all methods for all scenarios during training.}
    \label{fig:result}
\end{figure*}

 The average reward per time step, collision rate, and success rate across all training scenarios are depicted in Fig. \ref{fig:result}. 
 Table \ref{table:1} showcases the best scores achieved by each method across all training scenarios, providing a comprehensive overview of performance comparisons.

 Baseline 2 consistently outperformed baseline 1 in every scenario and key metric while demonstrating stable learning. The robustness of baseline 2 against action variance enabled it to explore and identify optimal states and actions, resulting in stable learning and improved performance. Furthermore, agents utilizing RP, IRP, and IRP with uncertainty propagation exhibit stable learning compared to the baseline methods, as illustrated in Fig. \ref{fig:result}. Across all scenarios, these agents consistently outperformed the baseline methods, showing sequential performance improvements.
The agent with RP outperformed the baseline methods, which is attributed to its utilization of the error reduction property of N-step SARSA and explicit learning of the consequences of actions.
 The agent with IRP surpassed the agent with RP due to accurate prediction of future actions, resulting in more precise reward prediction and improved overall performance. Additionally, the agent with IRP and uncertainty propagation outperformed the agent with IRP, benefiting from the consideration of uncertainty.

In scenario 3, although lane changes are quite challenging in typical road situations, the collision rate was relatively low compared to the other scenarios. This was due to the conservative driving behavior of the other traffic participants used in the experiment, which helped reduce the collision rate. In scenario 4, the collision rate was much higher than in scenarios 1-3 because it was considerably more challenging to avoid parked cars while also avoiding the other traffic participants, compared to the other scenarios. A video of the agent with IRP and uncertainty propagation is available at \url{https://www.youtube.com/watch?v=PfDbaeLfcN4}. 



\begin{table}[t!]
\centering
\captionsetup{size=footnotesize}
\caption{Best scores for each compared method for all scenarios during training.}
\begin{tabular}
{ |p{0.2cm} p{1.2cm}||p{1.5cm}|p{1.5cm}|p{1.5cm}|p{1.5cm}|}
 \hline
 Scenario & & Average& Collision& Success\\
 & &reward&rate&rate\\
 \hline
 1& Baseline 1& 0.0589 & 42.81\%& 20.65\%\\
  & Baseline 2& 0.0942 & 28.32\%& 70.95\%\\
  & RP& 0.1854 & 14.83\%& 84.15\%\\
  & IRP& 0.2194 & 13.47\%& 84.81\%\\
  & IRP+UP& \textbf{0.2297}& \textbf{5.682\%}& \textbf{93.86\%}\\
 \hline
 2& Baseline 1& 0.0383 & 61.71\%& 21.68\%\\
  & Baseline 2& 0.0912 & 43.04\%& 56.73\%\\
  & RP& 0.2200 & 16.75\%& 82.79\%\\
  & IRP& 0.2252 & 8.385\%& 91.50\%\\
  & IRP+UP& \textbf{0.2460} & \textbf{5.782\%}& \textbf{94.11\%}\\
 \hline
 3& Baseline 1& 0.0073 & 60.04\%& 15.81\%\\
  & Baseline 2& 0.1424 & 42.12\%& 19.62\%\\
  & RP& 0.1980 & 14.16\%& 85.86\%\\
  & IRP& 0.2083 & 5.913\%& 93.85\%\\
  & IRP+UP& \textbf{0.2250} & \textbf{3.590\%}& \textbf{96.08\%}\\
 \hline
 4& Baseline 1& 0.0694 & 78.43\%& 6.911\%\\
  & Baseline 2& 0.0909 & 54.65\%& 37.84\%\\
  & RP& 0.1870 & 31.87\%& 67.67\%\\
  & IRP& 0.2173 & 21.55\%& 78.40\%\\
  & IRP+UP& \textbf{0.2299}& \textbf{18.26\%}& \textbf{81.91\%}\\
  \hline  
\end{tabular}
\captionsetup{size=footnotesize}
\label{table:1}
\end{table}


%% file: 6_conclusion_and_future_work.tex
\section{Conclusions And Future Work} \label{conclusion_and_future_work}
In this study, methods including RP, IRP, and uncertainty propagation are proposed to reduce the function approximation error and improve the performance and learning stability of an AV's RL-based planning agent. The proposed method was evaluated under several scenarios, and the results demonstrated that the proposed method improves both learning stability and agent performance compared to baseline methods. However, the result of the evaluation showed that the proposed method still has poor performance in difficult and complex scenarios. Future work will involve increasing safety while having better performance.
